\documentclass[10pt,journal,compsoc]{IEEEtran}
\ifCLASSOPTIONcompsoc
  \usepackage[nocompress]{cite}
\else
  \usepackage{cite}
\fi

\ifCLASSINFOpdf

\else

\fi

\usepackage{amsmath}

\usepackage{algorithmic}

\usepackage{array}

\usepackage{url}

\usepackage{hyperref}
\usepackage{subfigure}
\usepackage{makeidx}
\usepackage{graphicx}
\usepackage{amsmath}
\usepackage{nccmath}
\usepackage{extarrows}
\usepackage{amsfonts,amsmath}
\usepackage{algorithm2e}
\usepackage{changepage}
\usepackage{float}
\usepackage{multirow}
\usepackage{amssymb}

\begin{document}

\title{Smooth Deformation Field-based\\ Mismatch Removal in Real-time}

\author{Haoyin Zhou,~\IEEEmembership{Member,~IEEE}
        and Jagadeesan Jayender,~\IEEEmembership{Senior Member,~IEEE}
\IEEEcompsocitemizethanks{\IEEEcompsocthanksitem
 Haoyin Zhou and Jagadeesan Jayender are with the Surgical Planning Laboratory, Brigham and Women's Hospital, Harvard Medical School, Boston,
 MA, 02115, USA.\protect\\
E-mail: zhouhaoyin@bwh.harvard.edu; jayender@bwh.harvard.edu.
}
\thanks{.}}

\markboth{Journal of \LaTeX\ Class Files,~Vol.~14, No.~8, August~2015}%
{Shell \MakeLowercase{\textit{et al.}}: Bare Demo of IEEEtran.cls for Computer Society Journals}

\IEEEtitleabstractindextext{%
\begin{abstract}

This paper studies the mismatch removal problem, which may serve as the subsequent step of feature matching. Non-rigid deformation makes it difficult to remove mismatches because no parametric transformation can be found. To solve this problem, we first propose an algorithm based on the re-weighting and 1-point RANSAC strategy (R1P-RNSC), which is a parametric method under a reasonable assumption that the non-rigid deformation can be approximately represented by multiple locally rigid transformations. R1P-RNSC is fast but suffers from a drawback that the local smoothing information cannot be taken into account. Then, we propose a non-parametric algorithm based on the expectation maximization algorithm and dual quaternion (EMDQ) representation to generate the smooth deformation field. The two algorithms compensate for the drawbacks of each other. Specifically, EMDQ needs good initial values provided by R1P-RNSC, and R1P-RNSC needs EMDQ for refinement. Experimental results with real-world data demonstrate that the combination of the two algorithms has the best accuracy compared to other state-of-the-art methods, which can handle up to 85\% of outliers in real-time. The ability to generate dense deformation field from sparse matches with outliers in real-time makes the proposed algorithms have many potential applications, such as non-rigid registration and SLAM.

\end{abstract}

}

\maketitle

\IEEEdisplaynontitleabstractindextext

\IEEEpeerreviewmaketitle

\IEEEraisesectionheading{\section{Introduction}\label{sec:introduction}}

\IEEEPARstart{F}{eature} matching is the foundation of many computer vision applications, such as simultaneously localization and mapping (SLAM) \cite{mur2017orb}, structure-from-motion (SfM) \cite{schonberger2016structure} and image registration \cite{sotiras2013deformable}. The problem of detecting key points and extracting their descriptors from the input images has been studied for decades and many effective methods have been proposed, such as SURF \cite{bay2006surf} and ORB \cite{rublee2011orb}. In recent years, deep learning-based methods have also been developed to improve the performance of feature detection and description \cite{yi2016lift}. Correspondences between these feature points can be built according to the computed descriptors \cite{muja2014scalable}. However, due to noise, illumination change, camera motion and object deformation, incorrect matches, or outliers, are unavoidable. To remove outliers from the correct matches is essential for the feature-based applications, and the related methods can be roughly classified into parametric and non-parametric methods.

The parametric methods assume that there exists a simple analytical transformation between two point clouds \cite{chin2016guaranteed}, which include affine \cite{lin2016shape}, homography \cite{zheng2018augmented}, epipolar geometry \cite{deriche1994robust} and so on. For example, Zheng et al proposed an algorithm to extract the planar homography matrix from the matches, which is fast and accurate but requires the observed objects to be approximately planar or the camera rotates at a fixed location \cite{zheng2018augmented}. To handle more general transformations, the most common method is random sample consensus (RANSAC) \cite{fischler1981random}, which has many modified versions \cite{chum2005matching}\cite{raguram2008comparative}\cite{chum2008optimal}. The RANSAC methods estimate the transformations from small sets of randomly selected control points, and select the transformation supported by the most number of matches as the final results. RANSAC is efficient in handling parametric transformations, but is difficult to handle non-parametric transformations, such as non-rigid deformation. To solve this problem, we propose a reasonable assumption that the non-rigid deformation can be approximately represented by multiple locally rigid transformations. However, traditional RANSAC methods have low efficiency under this assumption, because the selected sets of control points need to (1) be inliers and (2) can fit into the same rigid transformation. In this paper, we propose a novel re-weighting and 1-point RANSAC-based method (R1P-RNSC), which naturally satisfies the second condition because only one control point is needed in each trial. Hence, R1P-RNSC has high efficiency in handling non-rigid deformation while removing outliers in the feature matches.

Compared with parametric methods, non-parametric methods are more efficient in removing matching outliers when non-rigid deformation exists, which is an ill-posed problem. Most non-parametric methods are explicitly or implicitly based on the assumption that the deformation is smooth in local areas, and they recognize a match as an outlier if it is not consistent with its neighbors \cite{li2010rejecting}\cite{li2015pairwise}. For example, Probst et al used the isometric deformation prior that assumed the distances between matches are preserved under deformations \cite{probst2018model}. Ma et al proposed a method to estimate the vector field from the matches, which achieved high performance in terms of speed and accuracy \cite{ma2014robust}. They further proposed to use machine learning methods to exploit the consensus of local neighborhood structures \cite{ma2019lmr}. Hence, the representation of the local smoothing information is essential for the accuracy of the non-parametric methods. In this paper, we propose an algorithm called as EMDQ based on dual quaternion to generate the smooth deformation field. Dual quaternion \cite{kavan2007skinning} is widely used in the computer graphics field to generate smooth interpolations among 6-DoF rigid transformations. To reduce the impact of outliers, we integrate the dual quaternion-based interpolation process into the expectation maximization (EM) algorithm.

\begin{figure*} [htp]
\vspace{0.0cm}
\centering
  \includegraphics[width=1.0\textwidth]{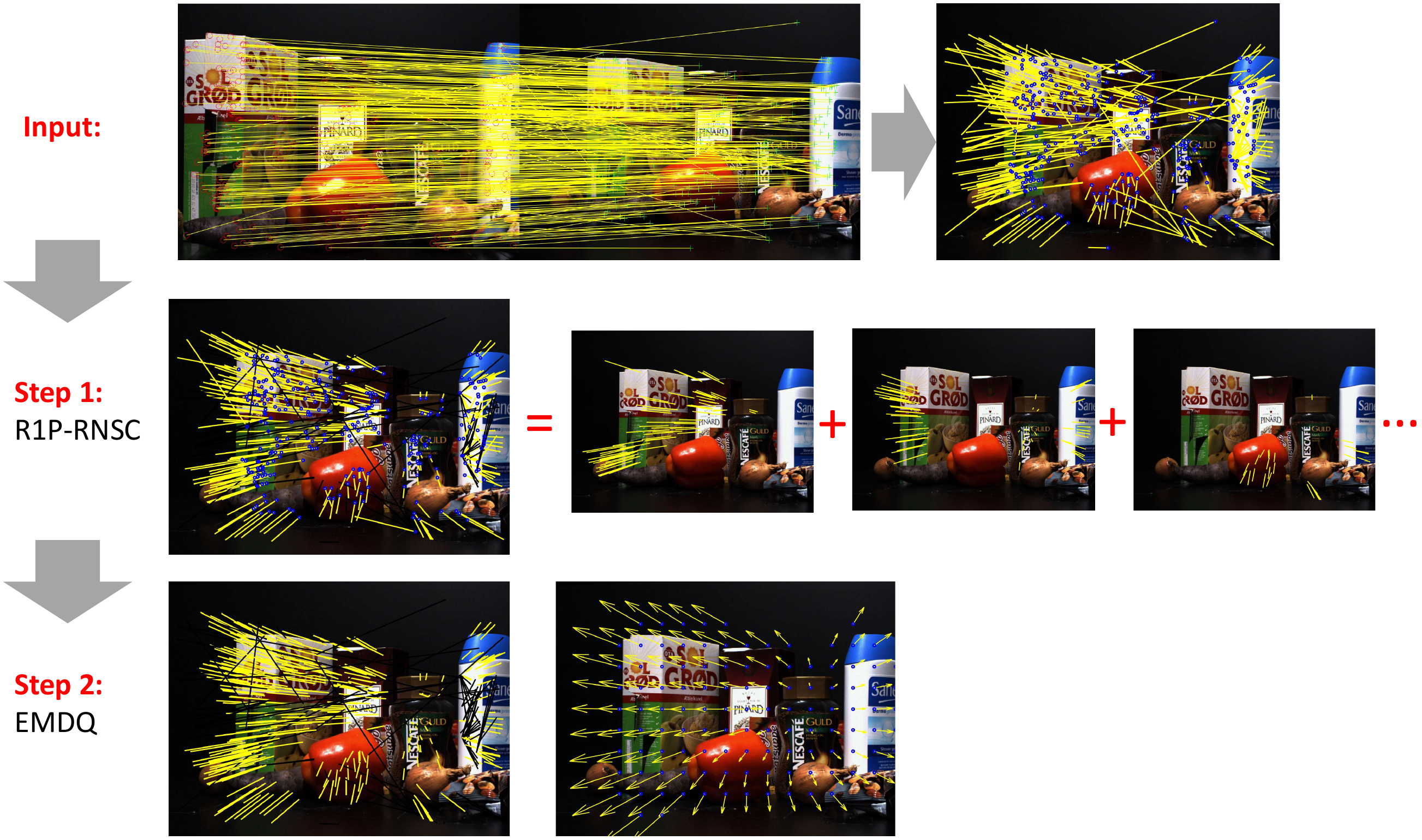}
  \caption{The process of the proposed method to remove outliers from feature matches. First row: The input images with matches built by traditional feature matching methods. Although the objects are rigid, the displacements of the correct matches in the images may be non-rigid. Second row: R1P-RNSC results, which assumes that the non-rigid deformation can be represented by multiple rigid transformations. Third row: EMDQ results and the generated smooth deformation field. In this example, the two steps took $\approx 20$ms to remove the outliers, and took additional $\approx 3$ms to generate the deformation field on the grid points.}
\label{fig_Process}
\end{figure*}

As shown in Fig.\ref{fig_Process}, the method proposed in this paper consists of two steps, which are R1P-RNSC and EMDQ. The two algorithms compensate for the drawbacks of each other. The accuracy of R1P-RNSC is limited due to the lack of ability to take into account smoothing information. However, R1P-RNSC is fast and able to extract the candidate rigid transformations from the matches, which is essential for the initialization of the EMDQ algorithm. The experimental results show that the combination of the two algorithms has the highest accuracy compared with other state-of-the-art methods, and the speed is sufficient for real-time applications.

This paper is organized as follows. In Section 2, we give the details of R1P-RNSC, including the re-weighting strategy and the termination conditions of the RANSAC framework. The details of the EMDQ algorithm are given in Section 3, in which we also provide the strategy of initialization based on the results of R1P-RNSC. Evaluation results are presented in Section 4. A discussion and description of planned future work is described in Section 5.

\subsection{Related Works}

Building correct correspondences between two point clouds in the presence of noise, outliers and deformation is a fundamental problem in the computer vision field, and the related research directions mainly include point clouds registration and feature matching \cite{maciel2003global}\cite{hacohen2011non}.

\noindent \textbf{(1) Point clouds registration:} This paper is closely related to the registration problem between point clouds or mesh models, which has been studied for decades with many effective solutions \cite{tam2012registration}. The goal of point clouds registration is to assign correspondences between two sets of points and/or to recover the transformation that maps one point set to the other. The most popular point clouds registration method is iterative closet points (ICP), which has many variants \cite{besl1992method}\cite{zhang1994iterative}\cite{segal2009generalized}. ICP iteratively assigns correspondences based on a closest distance criterion and finds the least-squares transformation between the two point clouds. To eliminate the impact of outliers, robust kernel functions can be applied \cite{newcombe2015dynamicfusion}. The ICP methods are sensitive to the initial alignment. To overcome this drawback, some works propose to use the soft-assignment strategy. For example, the robust point matching (RPM) algorithm \cite{chui2003new} uses a soft assignment matrix to represent the corresponding relationships. Another representative work is coherent point drift (CPD) \cite{myronenko2010point}, which is based on the expectation maximization (EM) algorithm to calculate the probabilities of the correspondences between points. CFD considers one point set as the Gaussian Mixture Model (GMM) centroids, and the other one as the data points. Then, CFD fits the GMM centroids to the data to maximize the likelihood. Inspired by CPD, we have also employed the EM algorithm to obtain the probabilities of the matches according to the error distances to reduce the impacts of outliers.

To provide initial values and improve the robustness of point cloud registration, some methods proposed to extract descriptors of key points from the point clouds and build correspondences \cite{deng2018ppfnet}. For example, Elbaz et al proposed a deep learning-based method called LORAX \cite{elbaz20173d} to describe the geometric structure of each point cloud with a low-dimensional descriptor. The obtained correspondences may include mismatches and an efficient method to remove these mismatches is needed.

\noindent \textbf{(2) Feature matching and outliers removal:} Feature matching is often a two-steps work. The first step detects feature points and builds matches. For example, there are many effective methods to detect key points and extract the descriptors from images, such as SIFT \cite{lowe2004distinctive}, SURF \cite{bay2006surf}, ORB \cite{rublee2011orb}, A-SIFT \cite{morel2009asift} and LIFT \cite{yi2016lift}. There are also many works that aim to develop more robust and efficient matchers, such as FLANN \cite{indyk1998approximate} \cite{bian2017gms}. However, mismatches are often unavoidable after the first step, hence a following step to remove the mismatches is necessary. The problems in the first step have been studied for many years but little attention has been given to identifying outliers in the obtained matches.

It is very common that the correct matches in the images have a non-rigid deformation, because even when observing a rigid three-dimensional (3D) object from different angles and positions, the displacements of different parts of the object on the two-dimensional (2D) images are often non-rigid. It is difficult to pre-define a transform model to handle the non-rigid deformation when the 3D structures of the object are unknown. Hence, non-parametric methods are more flexible in handling the non-rigid deformation. For example, Li et al proposed the ICF algorithm \cite{li2010rejecting} based on a diagnostic technique and SVM to learn correspondence functions that mutually map one point set to the other. Then mismatches are identified according to the consistence with the estimated correspondence functions. Lin et al. proposed the SIM algorithm \cite{lin2016shape} that achieves affine invariance by computing the shape interaction matrices of two corresponding point sets. SIM detects the mismatches by removing the most different entries between the interaction matrices. Basically, these methods interpolate a non-parametric function by applying the prior condition, in which the motion field associated with the feature correspondence is slow-and-smooth. Similarly, a recent work by Lin et al \cite{lin2017code} proposed a regression algorithm based on as-smooth-as-possible piece-wise constraints to discover coherence from the input matches.

Graph matching \cite{yan2015multi}\cite{cho2012mode} is another type of feature matching methods, which uses the nodes and edges to represent features and connections respectively. For example, Leordeanu et al proposed a spectral approach that uses weighted adjacency matrix to represent pairwise agreements between matches \cite{leordeanu2005spectral}. Torresani et al propose a graph matching optimization technique to minimize the energy function defined by the appearance and the spatial arrangement of the features \cite{torresani2008feature}. Liu et al proposed a graph-based method to handle multiple patterns \cite{liu2010common}. Cho et al proposed a method based on hierarchical agglomerative clustering \cite{cho2009feature}. The graph matching methods are flexible to multiple types of transformations but suffer from the NP-hard nature.

\section{Re-weighting and 1-point RANSAC (R1P-RNSC)}

Under a reasonable assumption that a non-rigid deformation can be approximatively represented by multiple locally rigid transformations, we propose a method based on the re-weighting and 1-point RANSAC (R1P-RNSC) strategy to handle matching outliers when non-rigid deformation exists. Compared with traditional RANSAC methods that are based three or four control points, the ability to use only one control point to estimate the rigid transformations can naturally satisfy the requirement that all control points should have similar rigid transformations. The R1P-RNSC method is fast and can provide good initial values for the subsequent refinement step.

\subsection{Re-weighting}

Matches between two point clouds may include outliers. Denote the two coordinates of the $i$th match as ${\bf{x}}_i \in \mathbb{R}^{D}$ and ${\bf{y}}_i \in \mathbb{R}^{D}$, $i=1,2,...N$, where $D=$ 2 or 3 is the dimension of the point clouds. The rigid transformation between ${\bf{x}}_i$ and ${\bf{y}}_i$ can be represented by

\begin{equation}
{\bf{y}}_i = \mu \left( {\bf{R}}{\bf{x}}_i + {\bf{t}} \right),
\label{eq_rigidtransform}
\end{equation}

\noindent where ${\bf{R}} \in SO(D)$ is the rotation matrix, ${\bf{t}} \in \mathbb{R}^{D}$ is the translation vector and $\mu \in \mathbb{R}$ is the scale factor.

In the R1PRNSC algorithm, we randomly select a match $o$ as the control point, then the coordinates of other matches with respect to match $o$ are

\begin{equation}
{\bf{y}}_i - {\bf{y}}_o = \mu {\bf{R}} \left( {\bf{x}}_i - {\bf{x}}_o \right), i = 1,2,...,N,
\label{eq_rigidtransform_o}
\end{equation}

\noindent which suggests that a rigid transform can be represented by $\bf{R}$, $\mu$, ${\bf{x}}_o$ and ${\bf{y}}_o$. Because ${\bf{x}}_o$ and ${\bf{y}}_o$ are constants, only $\bf{R}$ and $\mu$ are needed to be estimated. According to Ref.\cite{arun1987least}, $\bf{R}$ and $\mu$ can be obtained from ${\bf{x}}_i$ and ${\bf{y}}_i$, $i=1,2,...N$, by

\begin{equation}
[{\bf{U}},{\bf{\Sigma }},{{\bf{V}}^T}] = {\rm{svd}}({\bf{Y}}{{\bf{X}}^T}),{\bf{R}} = {\bf{U}}{{\bf{V}}^T},
\label{eq_R}
\end{equation}

\begin{equation}
\mu  = \left\| {{\rm{vector}}({\bf{Y}})} \right\|/\left\| {{\rm{vector}}({\bf{X}})} \right\|,
\label{eq_mu}
\end{equation}

\noindent where

\begin{equation}
{\bf{X}} = {\left[ {\begin{array}{*{20}{c}}
{{{\bf{x}}_1} - {{\bf{x}}_o}}& \cdots &{{{\bf{x}}_N} - {{\bf{x}}_o}}
\end{array}} \right]_{D \times N}},
\end{equation}

\begin{equation}
{\bf{Y}} = {\left[ {\begin{array}{*{20}{c}}
{{{\bf{y}}_1} - {{\bf{y}}_o}}& \cdots &{{{\bf{y}}_N} - {{\bf{y}}_o}}
\end{array}} \right]_{D \times N}}.
\end{equation}

Because match outliers exist and not all match inliers can fit into the local rigid transform of match $o$, the estimation of the rigid transformation represented by ${\bf{R}}$ and $\mu$ may be incorrect. We propose a re-weighting method to dynamically update the weights of matches $i = 1,2,...N$ by

\begin{equation}
{d_i} = \left\| {{{\bf{y}}_i} - {{\bf{y}}_o} - \mu {\bf{R}}\left( {{{\bf{x}}_i} - {{\bf{x}}_o}} \right)} \right\|,
\label{eq_di}
\end{equation}

\begin{equation}
{w_i} = \min \left( {H/{d_i},1} \right),
\label{eq_wi}
\end{equation}

\noindent where $d_i$ is the error distance of match $i$ with the estimated $\bf{R}$ and $\mu$, $H$ is a pre-defined threshold that when the error distance $d_i < H$, match $i$ is recognized as an inliers.

With the obtained weights of matches $w_i$, $i=1,2,...N$, the related items of matrices $\bf{X}$ and $\bf{Y}$ are adjusted according to the weights. Small weights $w_i$ will be assigned to the match inliers that cannot fit into the same rigid transformation of match $o$ and the match outliers, hence the estimation of $\bf{R}$ and $\mu$ in \eqref{eq_R} and \eqref{eq_mu} will not be interfered.

We perform the re-weighting strategy within an iterative process. With a randomly selected match $o$ as the control point, we set the initial values of weights as $w_i = 1$ for all matches. Then, we alternatively update $\bf{R}$ and $\mu$ according to \eqref{eq_R} and \eqref{eq_mu}, and update the weights of matches $w_i$,$i=1,2,...N$, according to \eqref{eq_di} and \eqref{eq_wi}. In practice we find this iteration process only need a small number of iterations to converge, and we choose the number of iterations 3 in the experiments.

At the end of the iteration process, we recover the translation $\bf{t}$ by

\begin{equation}
{\bf{t}} = {{\bf{y}}_o}/\mu  - {\bf{R}}{{\bf{x}}_o},
\label{eq_t}
\end{equation}

\noindent which is not needed by R1P-RNSC for removing outliers, but the subsequent EMDQ algorithm need the estimation results of rigid transformations represented by $\bf{R}$, $\bf{t}$ and $\mu$ to generate the initial dual quaternions.

\subsection{Modified RANSAC}

Traditional RANSAC methods aim to find the consensus that is supported by the most number of input data, and other consensuses are simply abandoned. It is difficult to use a single parametric transformation to represent the non-rigid deformation. To solve this problem, we propose to use multiple rigid transformations. In the R1P-RNSC algorithm, the rigid transformations are obtained in different RANSAC trials with different control points $o$. Hence, it is necessary to modify the standard RANSAC methods to reserve the obtained rigid transformations and design a new termination condition.

\subsubsection{Results Reservation:}
Within the RANSAC framework, we randomly try different matches as the control point $o$. With a control point $o$, some matches may have small error distances $d_i$ as in \eqref{eq_di} and can be considered as the candidate inliers. Denote the number of candidate inliers detected with control point $o$ as $T_o$, we omit the results if $T_o < T_{\rm{min}}$, where $T_{\rm{min}} \in {\mathbb{R}}$ is a threshold. A small $T_o$ suggests that the selected match $o$ may be an outlier and the results obtained from it is incorrect. If $T_o \ge T_{\rm{min}}$, we add the candidate inliers to the final inliers detection results, and update the inliers ratio by

\begin{equation}
\gamma  = {N_{{\rm{inliers}}}}/N,
\end{equation}

\noindent where $N_{\rm{inliers}}$ is the total number of detected inliers.

In our algorithm, the control point $o$ is only selected from the matches that are not considered as inliers so far, which aims to avoid repeatedly find the same rigid transformation.

\subsubsection{RANSAC Termination:}

The standard termination condition of RANSAC is determined by the number of trials and the largest number of detected inliers in one trial, which needs to be modified for the R1P-RNSC algorithm. The basic idea of the modified RANSAC termination condition is to terminate the RANSAC process when inliers are unlikely to exist in the remaining matches.

$T_{\rm{min}}$ is the minimal number of inliers detected by one trial that can be reserved. Assuming that there exist a set of $T_{\rm{min}}$ inliers in the remaining matches, then the possibility that a selected control point $o$ belongs to this set of inliers is $T_{\rm{min}} / (N-\gamma N)$. Then, the possibility that after $k$ trials of different control point $o$, the possibility that the a set of $T_{\rm{min}}$ inliers is not found is ${\left( {1 - {T_{{\rm{min}}}}/(N - \gamma N)} \right)^k}$, which equals to $1-p$. $p$ is the widely used RANSAC parameter \cite{wiki:RANSAC}, which suggests the desired probability that the RANSAC algorithm provides a useful result after running and in our algorithm we set $p = 0.95$. Hence, the termination of our 1-point RANSAC framework is

\begin{equation}
k > \frac{{\log (1 - p)}}{{\log (1 - {T_{{\rm{min}}}}/(N - \gamma N))}},
\end{equation}

\noindent where $k$ is the number of RANSAC trials.

\subsection{Sparse Implementation}

The R1P-RNSC algorithm computes the rigid transformations by re-weighting all matches. A more efficient way is to use a small number of sample points $N_{{\rm{sparse}}} < N$ to compute the rigid transformations, and then apply the obtained rigid transformations back to all matches to determine inliers and outliers. This sparse implementation is less accurate because it may miss possible rigid transformations to represent the non-rigid deformation, but is faster due the the smaller number of matches involved in the re-weighting process.

\subsection{Remaining Problem}

\begin{figure} [htp]
\vspace{0.0cm}
\centering
  \includegraphics[width=0.49\textwidth]{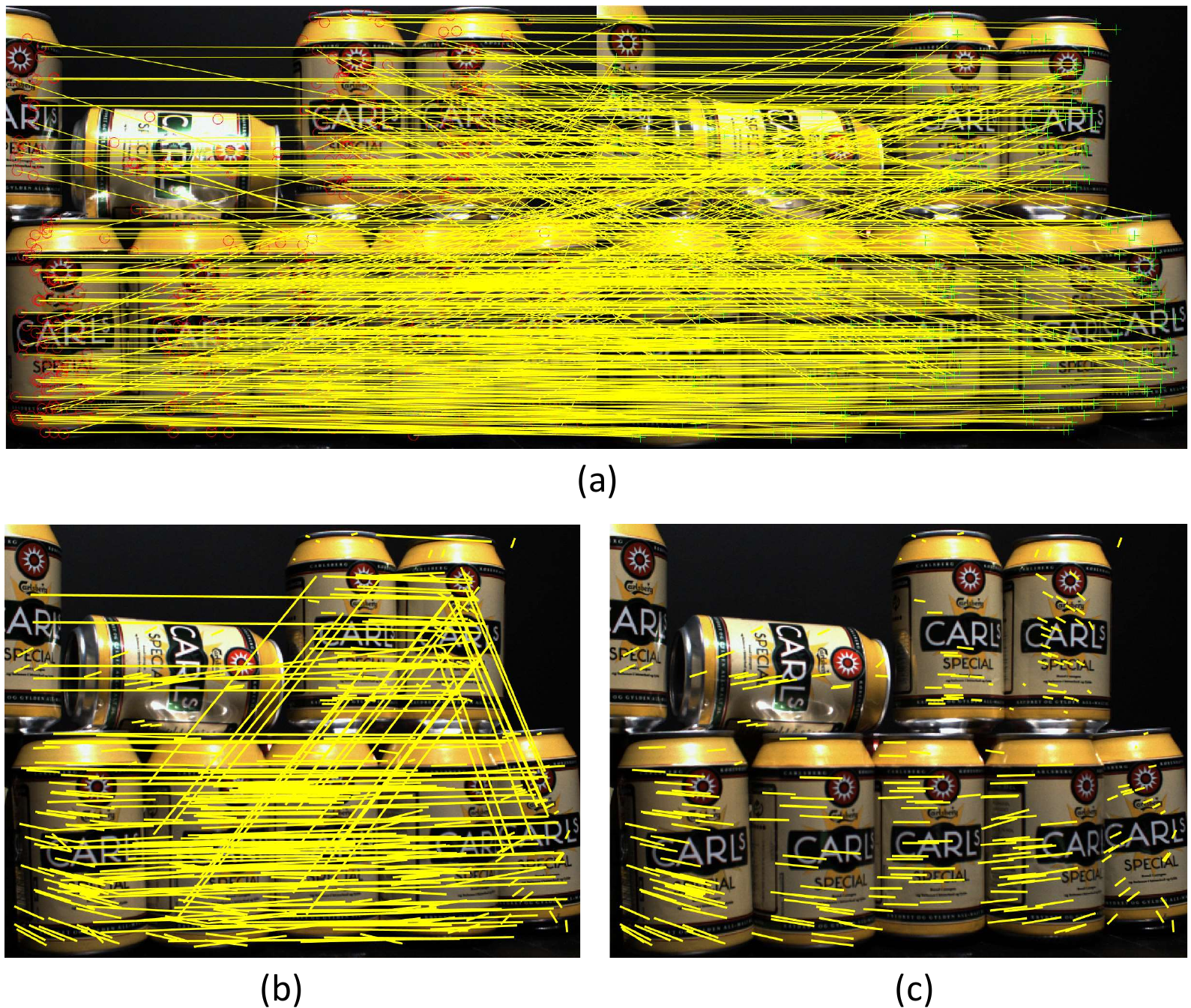}
\caption{The R1P-RNSC algorithm cannot take into account local smoothing information. A typical example is shown in this figure. (a) The repeating pattern lead to a large number of matching outliers, and (b) the results of the R1P-RNSC algorithm misclassified these matches as inliers and (c) The subsequent EMDQ step takes into account local smoothing information and obtains more accurate results.}
\label{fig_RepeatingPattern}
\end{figure}

The proposed R1P-RNSC algorithm is a parametric method that is efficient in estimating rigid transformations from the matches when outliers exist. However, R1P-RNSC suffers from a problem that it cannot take into account the smoothing information. For example, neighboring matches should have similar rigid transformations, but the estimated rigid transformations by R1P-RNSC are independent with different control points. Hence the accuracy is limited especially when repeating pattern exists. A typical example is shown in Fig. \ref{fig_RepeatingPattern}, the repeating pattern may cause the feature points of a pattern be incorrectly matched to another pattern. These matching outliers are consistent with each other and are recognized as inliers by R1P-RNSC, because the number of outliers that satisfy with the same rigid transformation may larger than the threshold $T_{\rm{min}}$. To solve this problem, we propose a non-parametric algorithm called as EMDQ as the refinement step.

\section{Expectation Maximization and Dual Quaternion (EMDQ)}

We propose a non-parametric algorithm called as EMDQ to handle matching outliers by generating the smooth deformation field. Denote a deformation field as ${{f}}$, $f({\bf{x}}_i)$ is the displacement of ${\bf{x}}_i$ caused by $f$. Matching inliers should be consistent with the deformation field $f$, which suggest

\begin{equation}
{\bf{y}}_i \approx {f}({\bf{x}}_i),
\label{eq_yifxi}
\end{equation}

\noindent for inliers.

According to \eqref{eq_yifxi}, the proposed EMDQ algorithm distinguishes inliers and outliers according to the error distances between ${\bf{y}}_i$ and ${{f}} ({\bf{x}}_i)$, $i=1,2,...N$. The ability to generate the smooth deformation field $f$ from the matches with outliers is essential, and we employ dual quaternion to solve this problem.

\subsection{Dual Quaternion-based Deformation Field}

We assign discrete transforms $g_i$, $i=1,2,...N$ to all matches, and the smooth deformation field $f$ is generated by interpolating among $g_i$. According to \eqref{eq_rigidtransform}, a transform $g_i$ consists of the rotation ${\bf{R}}_i$, the translation ${\bf{t}}_i$ and the scale $\mu_i$, which moves coordinate ${\bf{x}}_i$ to ${\bf{y}}_i$ by

\begin{equation}
{\bf{y}}_i = {g_i}({\bf{x}}_i) = \mu_i ({\bf{R}}_i {\bf{x}}_i + {\bf{t}}_i).
\label{eq_ygxiRt}
\end{equation}

Then, the value of $f$ at a location is equal to the weighted average value of neighboring $g_i$. In this outliers removal problem, we mainly focus on the values of $f$ at the locations of the matches ${\bf{x}}_i$, $i=1,2,...N$, that is

\begin{equation}
f_{{\bf{x}}_i} = \sum\limits_{j = 1}^N {w_j^i{g_j}},
\label{eq_fwg}
\end{equation}

\noindent where $w_j^i$ is the weight that suggests the impact from match $j$ to $i$. $f_{{\bf{x}}_i}$ is the value of $f$ at the coordinate ${\bf{x}}_i$. This weighted average strategy \eqref{eq_fwg} suggests that only when the transformations of neighboring matches are consistent with that of a match $i$, then match $i$ can be recognized as an inlier according to \eqref{eq_yifxi}.

However, when performing interpolation according to \eqref{eq_fwg}, the use of the rotation matrices and translation vectors as in \eqref{eq_ygxiRt} may lead to inaccurate results \cite{shoemake1985animating}. To overcome this difficulty, we introduce a mathematics tool called as dual quaternion (DQ) \cite{kavan2007skinning}.

A DQ is a 8- dimension vector that can represent a 3D rotation and a 3D translation simultaneously. A 2D transform can be considered as a 3D transform that is restricted at the $z=0$ plane, and 4 out of the 8 components of a DQ is always zero. To improve the efficiency, we only compute the 4 non-zero components in our EMDQ implementation when $D=2$. Hence, ${\bf{q}}_i$ is a 4- or 8- dimension vector for $D =$ 2 or 3 respectively. Denote a dual quaternion related to match $i$ as ${\bf{q}}_i$, we have

\begin{equation}
{\bf{y}}_i = {g_i}({\bf{x}}_i) = {\mu_i}{{\bf{q}}_i}({\bf{x}}_i),
\label{eq_qbt}
\end{equation}

\noindent where ${{\bf{q}}_i}({\bf{x}}_i)$ suggests to apply ${{\bf{q}}_i}$ to the coordinate ${\bf{x}}_i \in {\mathbb{R}}^{D}$ to obtain a transformed coordinate.

According to \eqref{eq_fwg} and \eqref{eq_qbt}, the deformation field $f$ at the location ${\bf{x}}_i$ is determined by the weighted average value of $\mu_j$ and ${\bf{q}}_j$, that is

\begin{equation}
f_{{\bf{x}}_i} = {\bar \mu}_i {\bar {\bf{q}}}_i
\end{equation}

\noindent where

\begin{equation}
{{\bf{\bar q}}_i} = \sum\limits_{j = 1}^N {\left( {w_j^i{{\bf{q}}_j}} \right)} /\left\| {\sum\limits_{j = 1}^N {\left( {w_j^i{{\bf{q}}_j}} \right)} } \right\|_{dq},
\label{eq_qimean}
\end{equation}

\begin{equation}
{{\bar \mu }_i} = \sum\limits_{j = 1}^N {\left( {w_j^i{\mu _j}} \right)} / {\sum\limits_{j = 1}^N {\left( {w_j^i} \right)} } ,
\label{eq_mumean}
\end{equation}

\noindent where $\left\| \cdot \right\|_{dq}$ is the norm of the dual quaternion.

Many works in the computer graphics field have proven the feasibility and efficiency of using the linear combination of DQ as in \eqref{eq_qimean} to generate smooth interpolations. For example, by using DQ-based interpolation from skeletons, the motion of the skin surface of a mesh model can be generated \cite{kavan2007skinning}. However, due to the existence of outliers, the interpolation process \eqref{eq_qimean} and \eqref{eq_mumean} may be incorrect. One way to reduce the impacts of outliers is to assign a small weight $w_j^i$ when match $j$ is an outlier. Hence, we employ the expectation maximization (EM) algorithm to dynamically update the weights $w_j^i$ in the EM iteration process.

\begin{figure*} [htp]
\vspace{0.0cm}
\centering
  \subfigure[]{
  \includegraphics[width=.40\textwidth]{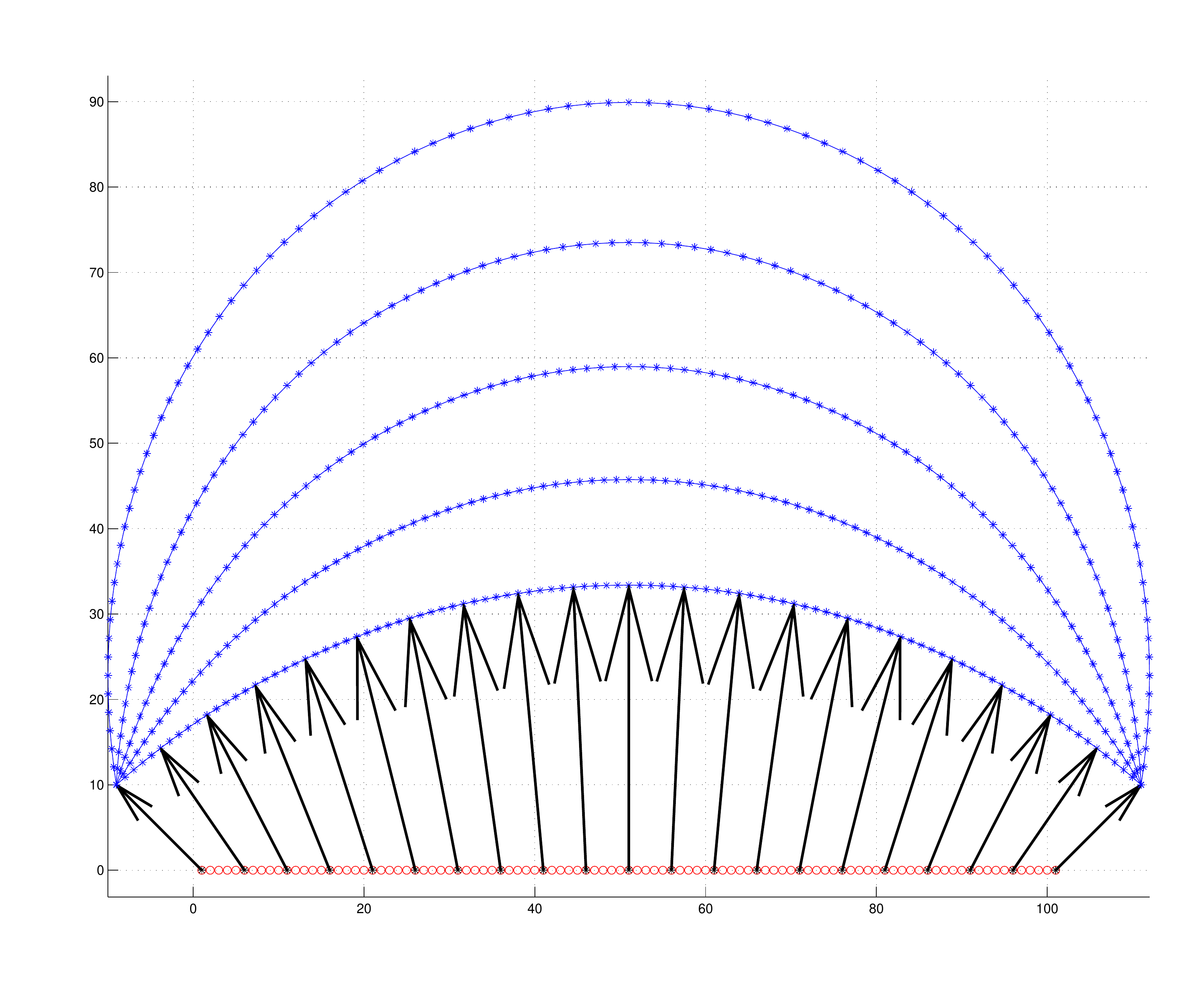}
  }
  \subfigure[]{
  \includegraphics[width=.40\textwidth]{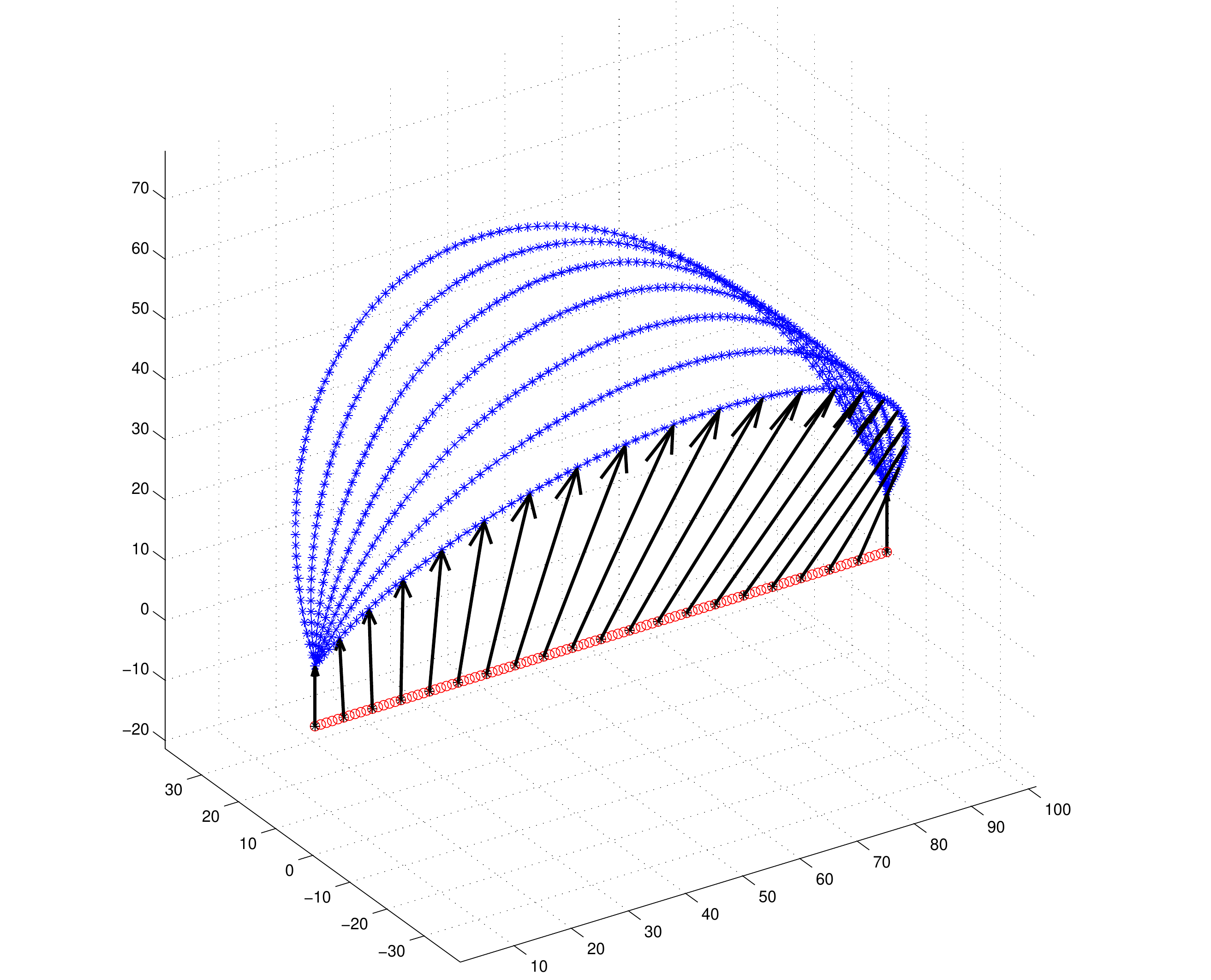}
  }
  \caption{Examples to show the uncertainty of smooth interpolation in (a) 2D and (b) 3D spaces respectively. The red and blue lines are the original and deformed point clouds respectively. The two end points are assigned with multiple different sets of dual quaternions, and other points are moved according to dual quaternion-based interpolation according to Eq.\eqref{eq_qimean} (in this example we keep the scale $\mu=1$). Although the sets of dual quaternion assigned to the end points result in the same displacements of the end points, the generated deformation of the point clouds may be different. All different deformations are smooth and may exist in reality, hence it is important to eliminate the uncertainty of interpolation and we use the results of R1P-RNSC to initialize the dual quaternion assigned to the matches.}
\label{fig_DQInterplation}
\end{figure*}

\subsection{EM Algorithm}

The EM algorithm is an iterative process that alternatively performs two steps. In the M-step, the probabilities that matches are inliers are computed, and the weights $w_j^i$ are further updated. In the E-step, the deformation field $f$ is updated according to the weights $w_j^i$. The details of the EM algorithm is as follows.

\noindent \textbf{(1) E-step:}

Inspired by \cite{myronenko2010point}, we consider the outliers removal problem as a classification problem according to the error distances between ${\bf{y}}_i$ and $f({\bf{x}}_i)$,$i=1,2,...N$. With a deformation field ${{f}}$, we have

\begin{equation}
p(i|{\rm{inlier}}) = \frac{1}{{2\pi {\sigma ^2}}}\exp (\frac{{{{\left\| {{{\bf{y}}_i} - {{f}({\bf{x}}_i)}} \right\|}^2}}}{{2{\sigma ^2}}})
\end{equation}

\noindent where $\sigma$ is the standard variation.

We consider the outliers follow an uniform distribution, that is
\begin{equation}
p(i|{\rm{oulier}}) = a,
\end{equation}

\noindent where $a$ is a constant.

Denote $\gamma$ as the inliers ratio, we have $p({\rm{inlier}}) = \gamma$ and $p(\rm{outlier})=1-\gamma$. According to Bayes' theorem, we have

\begin{equation}
\begin{aligned}
 &p({\rm{inlier}}|i) \\
 &= \frac{{p(i|{\rm{inlier}})p({\rm{inlier}})}}{{p(i|{\rm{inlier}})p({\rm{inlier}}) + p(i|{\rm{outlier}})p({\rm{outlier}})}} \\
 &= \frac{{\exp ({{\left\| {{{\bf{y}}_i} - f({{\bf{x}}_i})} \right\|}^2}/2{\sigma ^2})}}{{\exp ({{\left\| {{{\bf{y}}_i} - f({{\bf{x}}_i})} \right\|}^2}/2{\sigma ^2}) + 2\pi {\sigma ^2}\frac{{(1 - \gamma )}}{\gamma }a}}
 \end{aligned}
\label{eq_pinliersi}
\end{equation}

$p({\rm{inlier}}|i)$ suggests the probability that match $i$ is an inliers. We adjust the weight $w_j^i$, which is the influence of match $j$ to match $i$, according to $p({\rm{inlier}}|j)$. Taking into account the distances between matches, $w_j^i$ is determined by two parts, that is

\begin{equation}
w_j^i = w_{j,{\rm{distance}}}^i p({\rm{inliers}}|j) ,
\label{eq_wji}
\end{equation}

\noindent where $w_{j,{\rm{distance}}}^i$ is determined by the distance between match $i$ and $j$, that is

\begin{equation}
w_{j,{\rm{distance}}}^i = \max \left( {\exp (\frac{{{{\left\| {{{\bf{y}}_i} - {{\bf{y}}_j}} \right\|}^2}}}{{2{r^2}}}),\exp (\frac{{{{\left\| {{{\bf{x}}_i} - {{\bf{x}}_j}} \right\|}^2}}}{{2{r^2}}})} \right)
\end{equation}

\noindent where $r \in \mathbb{R}$ is a pre-defined radius. Because $w_{j,{\rm{distance}}}^i$ does not contain variables, we only compute it at the initial phase.

\noindent \textbf{(2) M-step:}

According to the updated weight $w_j^i$, we update ${{\bf{\bar q}}_i}$ and ${{{\bar \mu}}_i}$ for each match $i$ according to \eqref{eq_qimean} and \eqref{eq_mumean} respectively, which represent the value of the deformation field ${f}$ at ${\bf{x}}_i$, or ${f}_{{\bf{x}}_i}$.

The standard variation $\sigma$ is updated by

\begin{equation}
{\sigma ^2} = \sum\limits_{i = 1}^N {\left( {p({\rm{inlier}}|i){{\left\| {{{\bf{y}}_i} - {{ f}({\bf{x}}_i)}} \right\|}^2}} \right)} /\sum\limits_{i = 1}^N {p({\rm{inlier}}|i)}.
\end{equation}

Finally, we update ${\bf{q}}_i$ and $\mu_i$ to keep the transformation $g_i$ close to the deformation field ${f}_{{\bf{x}}_i}$ by

\begin{equation}
\Delta {{\bf{q}}_i} = {\rm{trans2dq}}(({{\bf{y}}_i} - {f}({\bf{x}}_i))/{\bar \mu _i}),
\label{eq_em_updateq}
\end{equation}

\begin{equation}
{{\bf{q}}_{i,{\rm{new}}}} = {{\bf{q}}_{i,{\rm{old}}}}\Delta {{\bf{q}}_i},
\end{equation}

\begin{equation}
\mu_{i,{\rm{new}}} = {\bar \mu}_i.
\label{eq_em_updatemu}
\end{equation}

Equations \eqref{eq_em_updateq}-\eqref{eq_em_updatemu} aim to keep \eqref{eq_qbt} stands and make the transforms $g_i$ more smooth, and $\rm{trans2dq}$ represents the conversion of the transform to the DQ representation.

The termination condition of the EM algorithm determined by the change ${p{{({\rm{inlier}}|i)}}}$, that is

\begin{equation}
\frac{1}{N}\left( {\sum\limits_{i = 1}^N {\left| {p{{({\rm{inlier}}|i)}^{(k)}} - p{{({\rm{inlier}}|i)}^{(k - 1)}}} \right|} } \right) < \theta
\end{equation}

\noindent where $\theta$ is a threshold. $k$ suggests the value of the $k$th iteration in the EM algorithm. After termination, matches with ${p{{({\rm{inlier}}|i)}}} > p_{\rm{min}}$ and ${\bf{y}}_i - f({\bf{x}}_i) < H$ are considered as inliers, where $p_{\rm{min}}$ is a threshold and $H$ is the same threshold used in the R1P-RNSC algorithm.

\subsection{Initialization from R1P-RNSC Results}

There are mainly three variables in the EMDQ algorithm that need to be initialized, which include the DQ ${\bf{q}}_i$, the scale factor $\mu_i$ and the weights between matches $w_j^i$. According to \eqref{eq_wji}, $w_j^i$ is determined by $w_{j,{\rm{distance}}}^i$ and $p({\rm{inliers}}|j)$, and $w_{j,{\rm{distance}}}^i$ is constant in the EM iterations. The initial values of the three variables determine the initial deformation field $f$ according to \eqref{eq_fwg}, and we use the results of R1P-RNSC for the initialization.

The proposed R1P-RNSC algorithm detects multiple rigid transformations. If a match $i$ is considered as an inlier with a rigid transformation detected by R1P-RNSC, then the related ${\bf{q}}_i$ and $\mu_i$ will be initialized according to the results. A match $i$ may satisfy multiple rigid transformations, and we use the one that has the most number of matches for EMDQ initialization. This initialization is important because as the examples shown in Fig. \ref{fig_DQInterplation}, there may exist multiple smooth deformations between two matches. Hence, if the ${\bf{q}}_i$ and $\mu_i$, $i=1,2,...N$ are randomly initialized, the deformation field $f$ may vary because it is generated by interpolating the ${\bf{q}}_i$ and $\mu_i$ of neighboring matches. Hence, to remove this uncertainty, EMDQ relies on the results of R1P-RNSC for the initialization of ${\bf{q}}_i$ and $\mu_i$, $i=1,2,...N$.

The initial values of $p({\rm{inliers}}|i)$ are important for reducing the impact of outliers when generating the initial deformation field $f$. The natural idea is to initialize $p({\rm{inliers}}|i) = 1$ if match $i$ is recognized as an inlier by R1P-RNSC, or $p({\rm{inliers}}|i) = 0$. However, we found that this initialization cannot handle the repeating pattern (see Fig. \ref{fig_RepeatingPattern}) very well, because at some local areas, there may be more false positives than true positives in the R1P-RNSC results, and the generated deformation field $f$ yields false positives. We propose a simple method to overcome this problem according to the uncertainty of the detected rigid transformations provided by R1P-RNSC. With match $o$ as the control point, R1P-RNSC may find different number of inliers, which is denoted as $T_o$. It is intuitive that when $T_o$ is large, the results are more reliable. Hence, we initialize $p({\rm{inliers}}|i) = T_o$ if match $i$ is recognized as an inlier with match $o$ as the control point. This allows that $f$ is initially generated from the most reliable matches.

\section{Experiments}

The performance of the proposed algorithms was evaluated against state-of-the-art matching outliers removal methods. The source code was implemented in MATLAB and executed on a computer with an Intel Core i7 2.60 GHz CPU. In our MATLAB implementation, we used the vlfeat open source library \cite{vedaldi2010vlfeat} for generating the kd-tree to obtain the neighboring matches.

The EMDQ algorithm requires R1P-RNSC for initialization. Hence, in this section, EMDQ reports the final accuracy and runtime results of the R1P-RNSC and EMDQ algorithms combined, while the sEMDQ reports the results when using the sparse implementation of R1P-RNSC.

\begin{figure*} [htp]
\vspace{0.0cm}
\centering
  \includegraphics[width=0.95\textwidth]{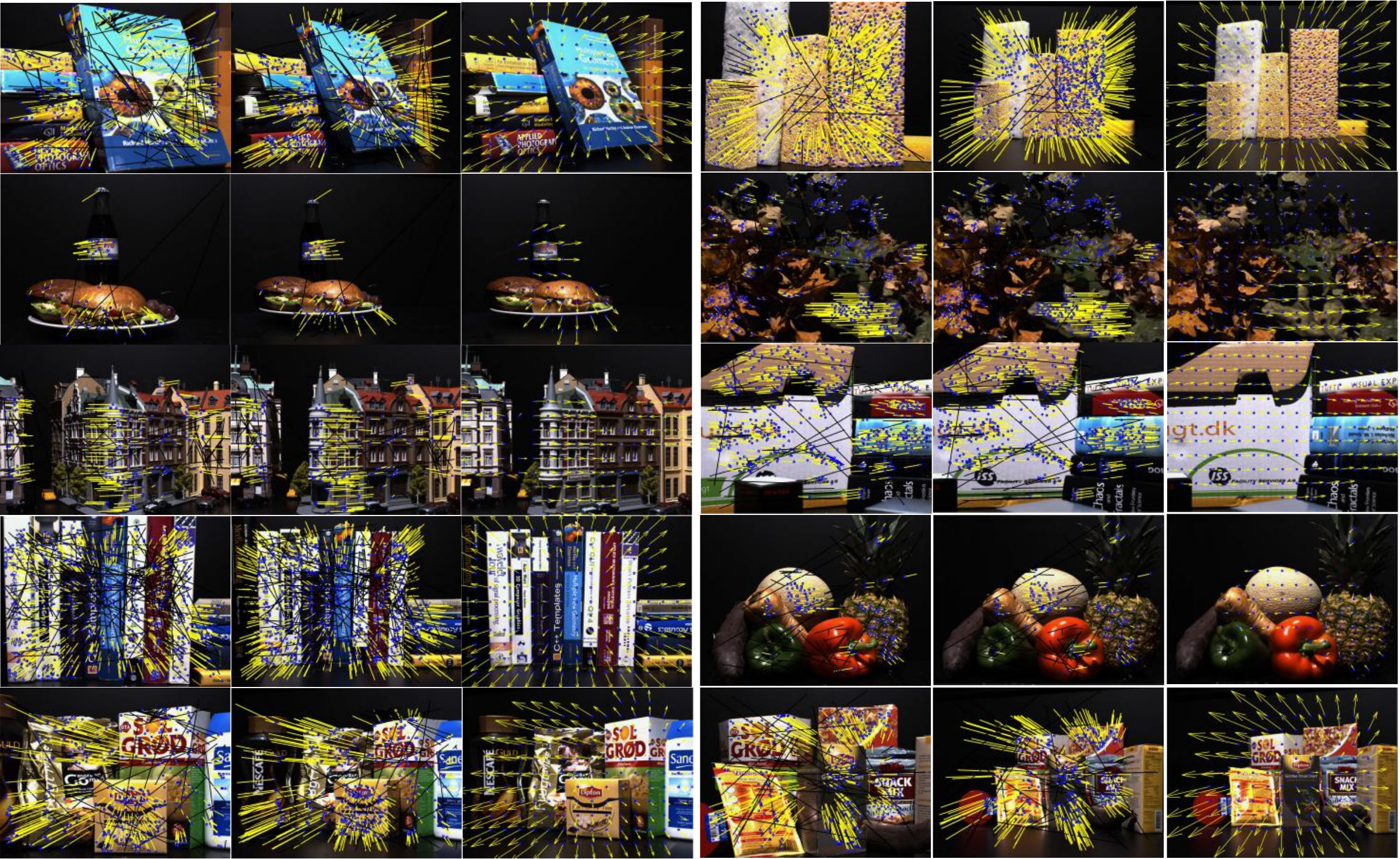}
\caption{Samples of the results of EMDQ on 2D DTU images. First and second columns: the two image pairs with SURF features, the yellow and black lines are identified as inliers and outliers by EMDQ respectively. Third column: the generated deformation field on grid sample points, which are obtained by taking the weighted average of neighboring $\bf{q}$ and $\mu$ of the match inliers. We only show the deformation field at locations that are close to match inliers, which suggests lower uncertainty. The grid points were sampled with a step of 50 pixels, and the generation of the deformation field on the grid points took less than 5 ms.}
\label{fig_DTU2DSamples}
\end{figure*}

\begin{figure*} [htp]
\vspace{0.0cm}
\centering
  \subfigure[]{
  \includegraphics[width=.48\textwidth]{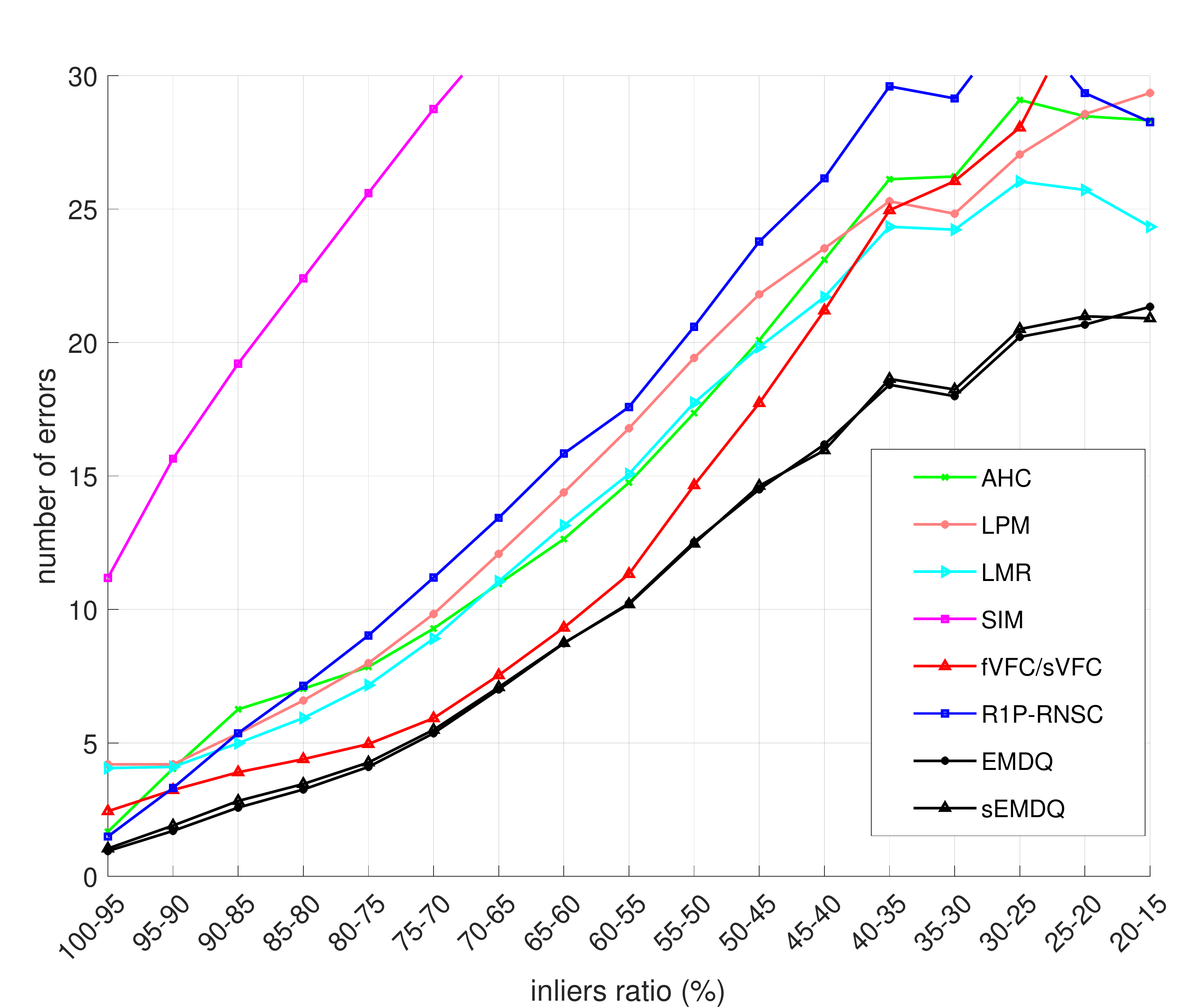}
  }
  \subfigure[]{
  \includegraphics[width=.48\textwidth]{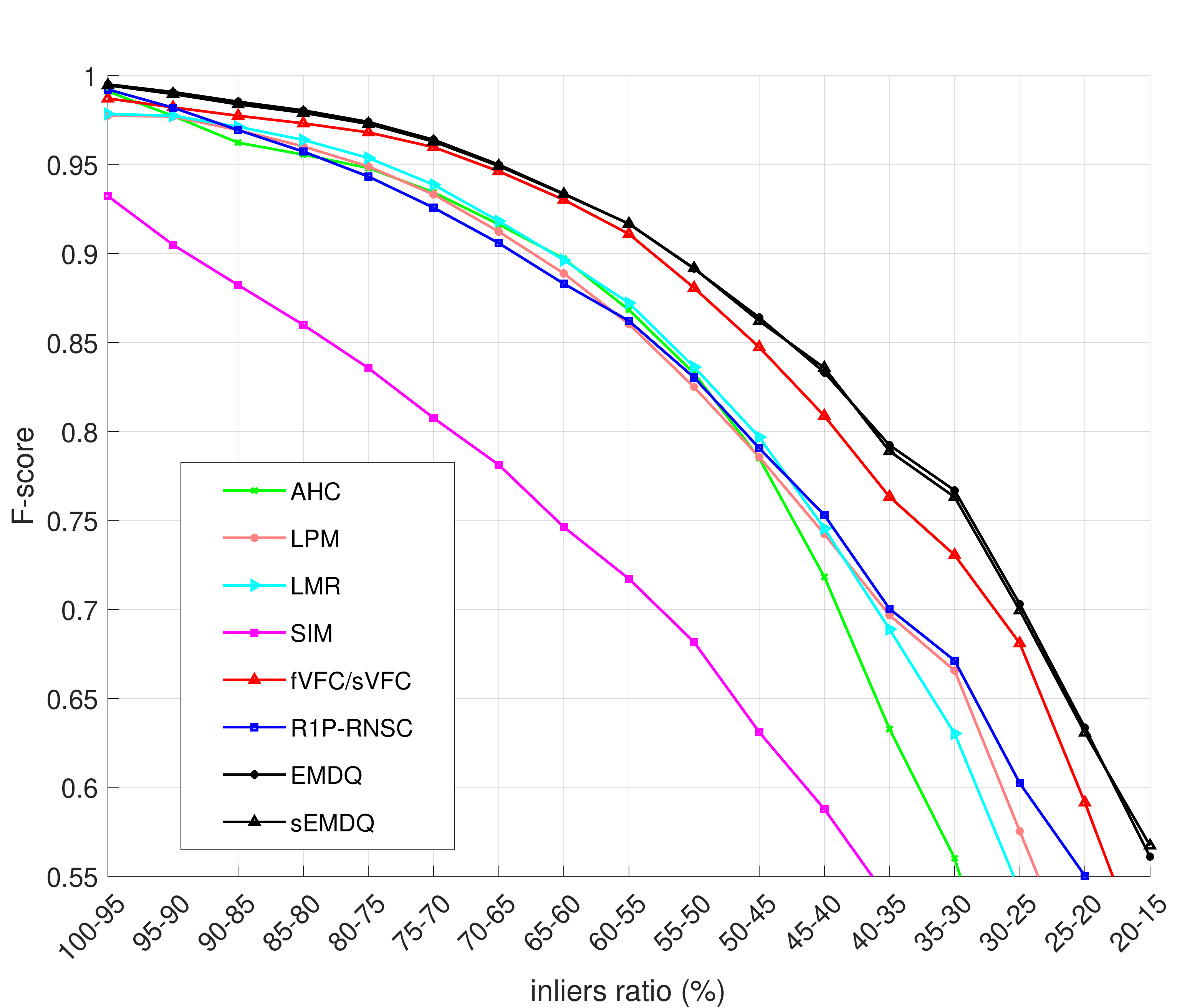}
  }
  \subfigure[]{
  \includegraphics[width=.48\textwidth]{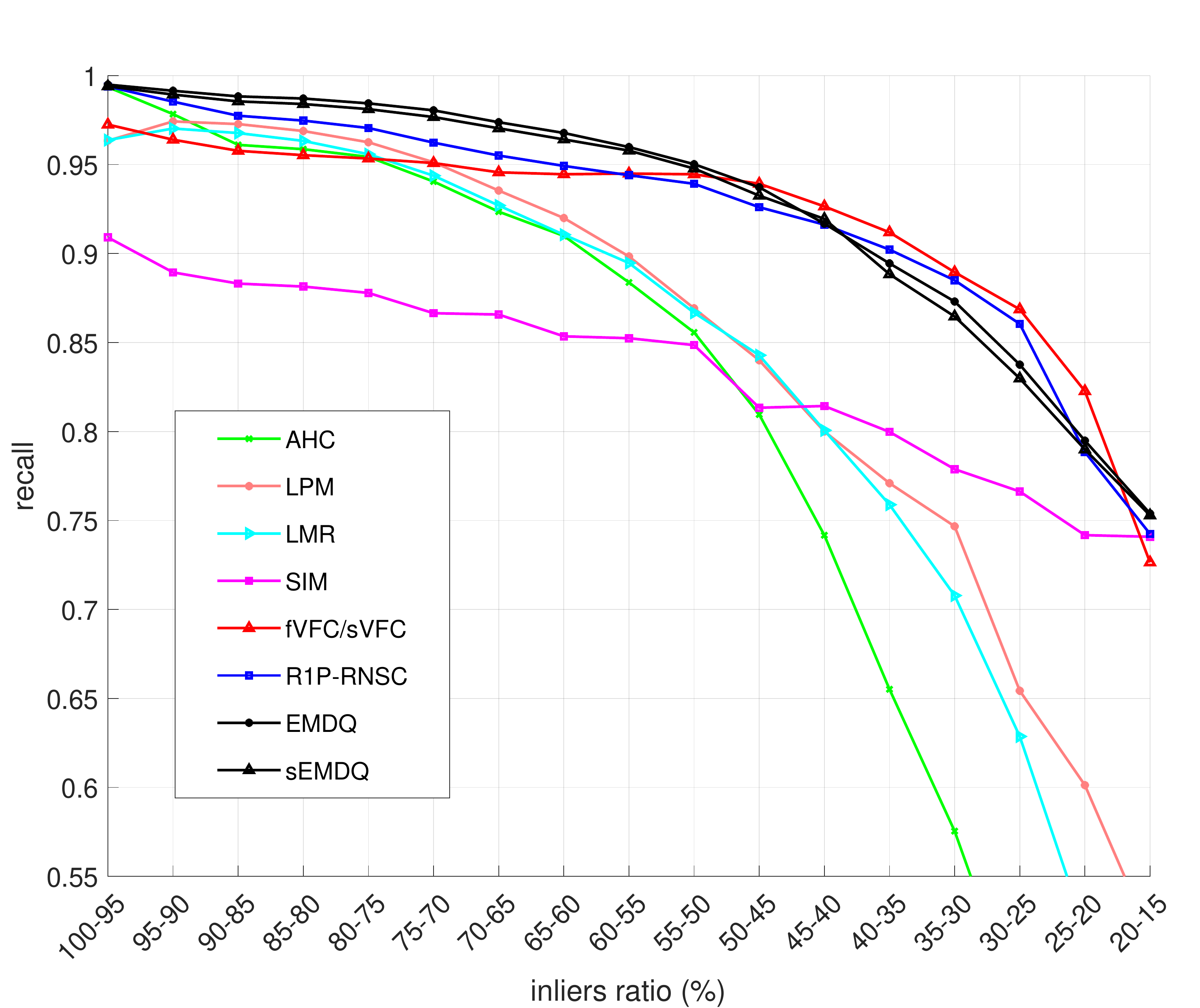}
  }
  \subfigure[]{
  \includegraphics[width=.48\textwidth]{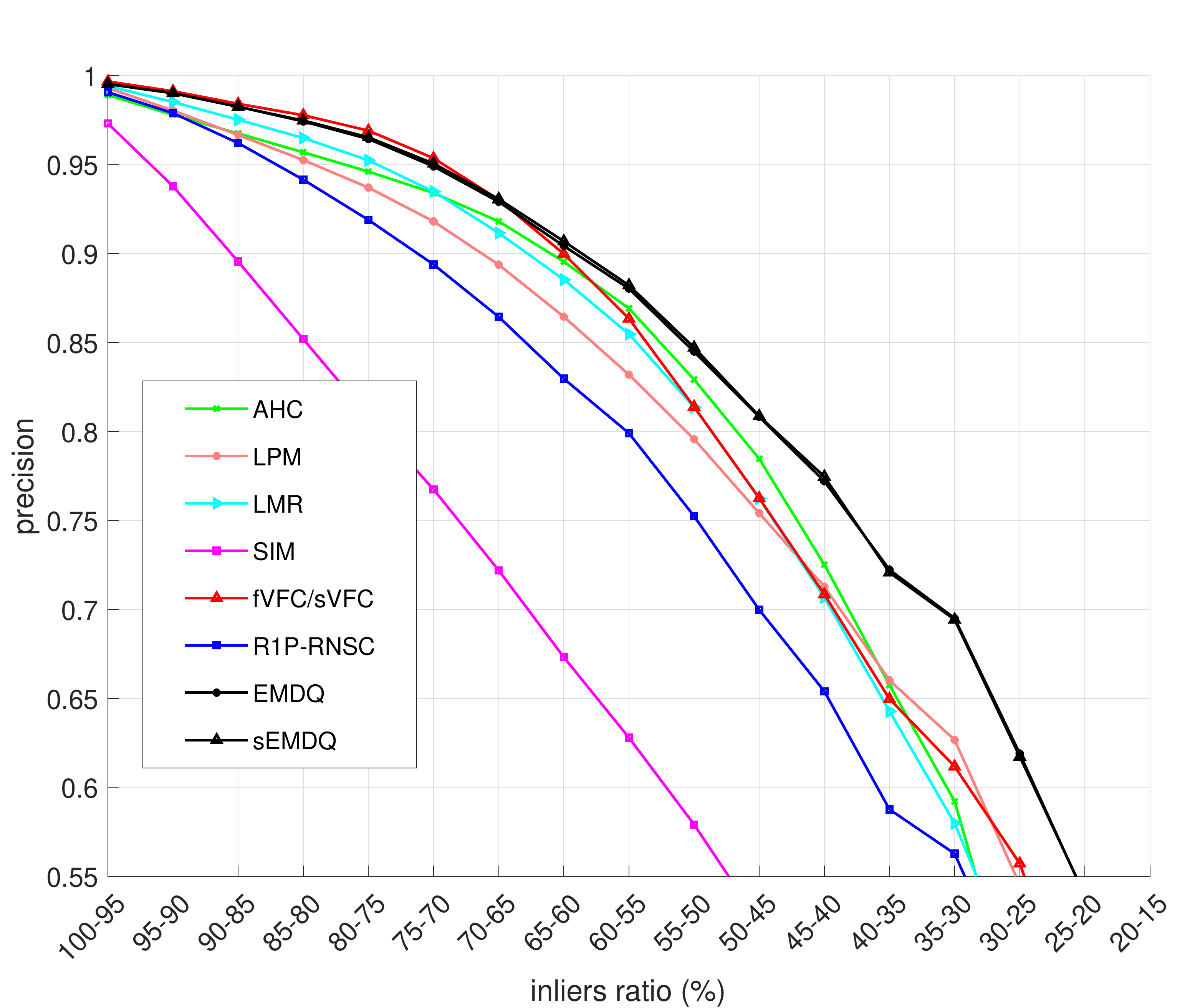}
  }
  \caption{Accuracy results with DTU data (2D experiments). (a) Number of errors. (b) $F$-$score$. (c) $Recall$. (d) $Precision$.}
\label{fig_DTU2DAccuracy}
\end{figure*}

\subsection{2D Experiments}

For 2D cases, the parameters of the R1P-RNSC algorithm are as follows: $H = 20$ pixels is the inliers threshold; $T_{{\rm{min}}} = 5$ is the least number of inliers found by one R1P-RNSC trial that will be added to the final R1P-RNSC results. We used a large $H$ and a small $T_{{\rm{min}}}$, which aims to find as many rigid transformations as possible to represent the non-rigid deformation. Many outliers may be mis-recognized as inliers by R1P-RNSC, and we mainly rely on the subsequent EMDQ algorithm to remove these mis-recognized outliers. The parameters of the EMDQ algorithms are as follows: $r = 50$ pixels is the radius to compute $w_{j,{\rm{distance}}}^i$ between matches; $a=1e{\rm{-}}5$ is used for computing $p(i|{\rm{outlier}})$; $p_{\rm{min}} = 0.5$ is the threshold for EMDQ to determine inliers and outliers; $\theta = 0.005$ is threshold used in the termination condition of EMDQ iterations; $N_{\rm{neighbor}} = 16$ is the number of neighboring matches that are considered in the EMDQ algorithm, which are found by kd-tree.

The 2D experiments were mainly conducted on the DTU robot image data \footnote{http://roboimagedata.compute.dtu.dk/} \cite{aanaes2012interesting}, which provides images and the related 3D point cloud obtained by structured light scan. The camera was fully calibrated and the true values of the camera positions and orientations are known. Images have a resolution of $800\times600$. Datasets numbered 1 to 30 were used. In each dataset, images were captured under 19 different illumination situations and from 119 camera positions. We randomly selected 10 out of 19 illumination situations. Hence, a total of $30\times10\times119 = 35700$ images were included in this evaluation. Following the instruction, for each dataset and illumination situation, we used the image numbered 25 as the reference image and performed SURF matching between the reference image and other images. By re-projecting the 3D point clouds to the images according to the true values of camera positions and orientations, we consider matches that have a re-projection error smaller than 10 pixels as inliers.

The outliers handling methods for comparisons include AHC \cite{zheng2018augmented}, LPM \cite{ma2019locality}, LMR \cite{ma2019lmr}, SIM \cite{lin2016shape} and VFC \cite{ma2014robust}. All the introduced methods were proposed in recent years and can reflect the state-of-the-art performances. It is worth noting that we did not introduce the traditional RANSAC + 3 or 4 control points methods for comparison because they cannot handle the non-rigid cases, which is the main focus of this paper. In our experiments, we did not change the parameters of these methods. VFC has a fast version and a sparse version, which are referred to as fVFC and sVFC in this section. AHC is a very fast method for removing the matching outliers, which focuses on $3 \times 3$ projective transformations, namely planar homographies. However, for the DTU dataset, the images were captured with rotating 3D objects, hence the planar homographies assumption may not stand. LMR is based on machine learning to solving the two-class classification problem, and we use the trained neural network model provided by the authors as the classifier in this experiments.

As shown in Fig.\ref{fig_DTU2DSamples}, the total number of matches and the percentage of outliers varied with objects, illumination situations and camera poses. Although clear comparisons require that only one factor be different, this kind of variable-controlling is difficult for the evaluation on real-world data because SURF matching results are unpredictable. In experiments we found that the accuracy of outliers removal algorithms was mainly affected by the percentage of outliers, and the runtime was mainly affected by the total number of matches. Therefore, in this section, we report the accuracy and runtime evaluation results by comparing the statistical results at each percentage of outliers and range of total number of matches respectively.

For accuracy evaluation, we report four metrics, which include the number of errors, $F$-$score$, $recall$ and $precision$. The number of errors suggests the total number of differences between the ground truth and the inliers-outliers recognition results. $F$-$score$ is widely used for statistical analysis of binary classification, and $F$-$score = 2 Recall \cdot Precision / (Recall + Precision)$. As shown in Fig. \ref{fig_DTU2DAccuracy}, the proposed EMDQ method has the best accuracy in terms of number of errors and the $F$-$score$. sEMDQ did not show an obvious decrease in accuracy compared with EMDQ. Because we set a non-restrict threshold for 1P-RNSC, the $recall$ rate of 1P-RNSC was high but the $precision$ rate was low, which results in a comparable F-score with AHC, LPM and LMR. fVFC and sVFC had very similar results hence we combined them in Fig. \ref{fig_DTU2DAccuracy}. When the inliers ratio was high, the $recall$ rate of fVFC/sVFC was lower than EMDQ/sEMDQ but the $precision$ rate was slightly higher. As the inliers ratio decreased, the $precision$ rate of fVFC/sVFC dropped significantly, which decreased their $F$-$score$ significantly although the $recall$ rate remained high.

\begin{figure} [htp]
\vspace{0.0cm}
\centering
  \includegraphics[width=.48\textwidth]{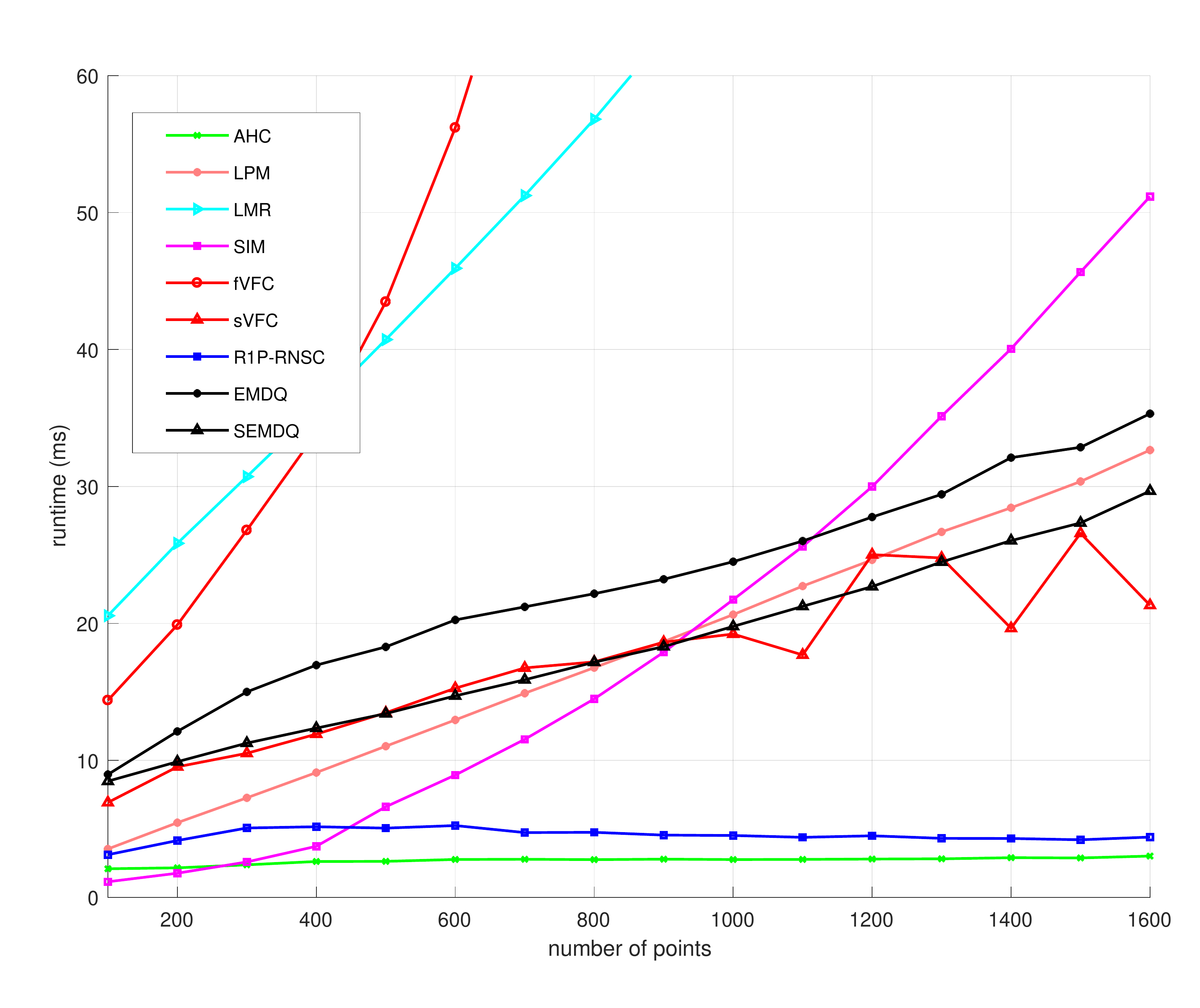}
  \caption{Runtime results with DTU data (2D experiments).}
\label{fig_DTU2DRuntime}
\end{figure}

As shown in Fig. \ref{fig_DTU2DRuntime}, sEMDQ had a comparable runtime with sVFC. AHC was fastest because it is based on an assumption that the object is planar, namely the planar homographies problem. This assumption makes AHC not suitable for solving the general outliers removal problem. Hence as shown in Fig. \ref{fig_DTU2DAccuracy}, the accuracy of AHC was lower than EMDQ and VFC. R1P-RNSC was also very fast.

\begin{figure*} [htp]
\vspace{0.0cm}
\centering
  \includegraphics[width=1.\textwidth]{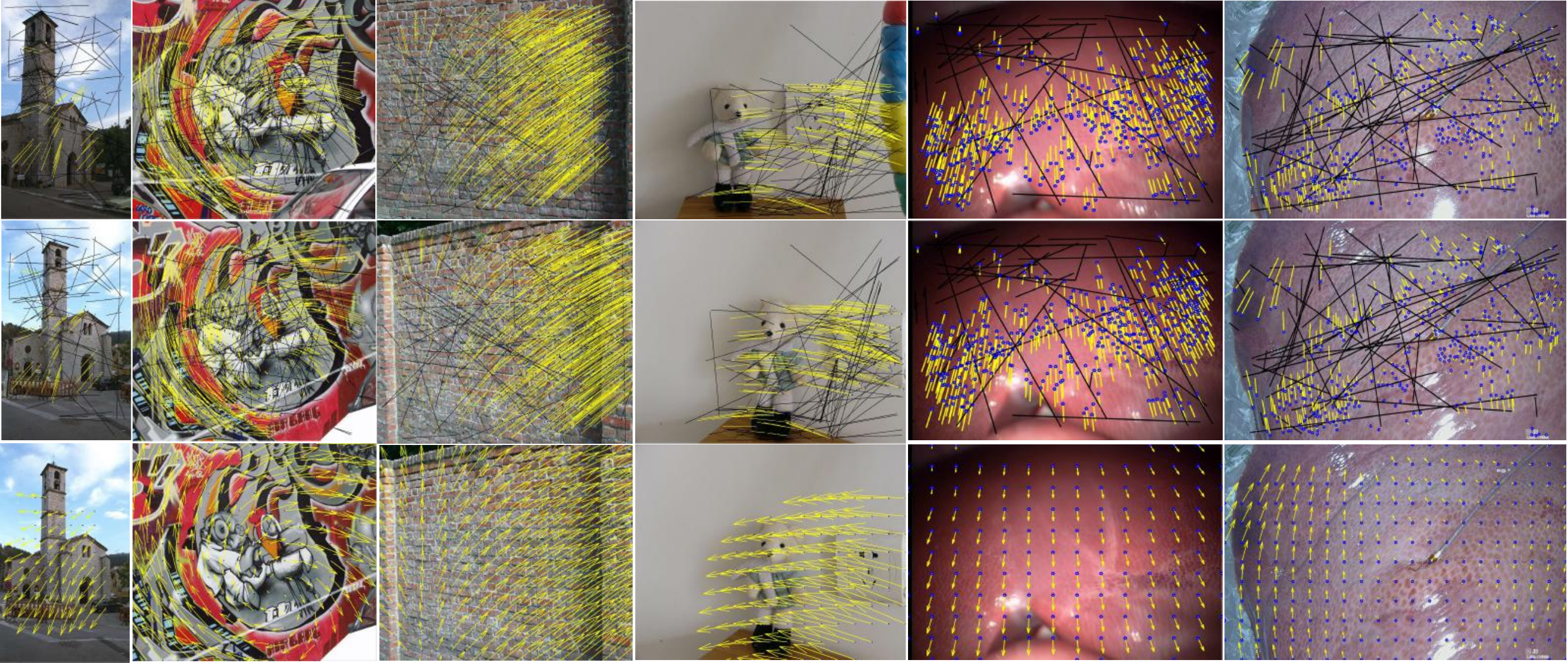}
\caption{Qualitative 2D experiments results for the EMDQ. First and second rows: the two image pairs with SURF features, the yellow and black lines are identified as inliers and outliers by EMDQ respectively. Third row: the generated deformation field. The last two data are soft tissues with non-rigid deformation. }
\label{fig_2DMedicalImages}
\end{figure*}

The DTU dataset contains large data with different illuminations and object motion, hence it is sufficient for the evaluation of the proposed algorithms. Although the objects in the DTU dataset are rigid, the displacements of corresponding feature points on the images have large non-rigid deformation, as shown in Fig. \ref{fig_DTU2DSamples}. In addition to the experiments on the DTU dataset, we also conducted qualitative experiments as shown in Fig. \ref{fig_2DMedicalImages}, which include laparoscopic images of soft tissue undergoing non-rigid deformation due to physiological motions such as breathing, and tissue manipulation. The generated deformation fields by EMDQ are smooth and physically realistic.

\subsection{3D Experiments}

Most parameters for 3D cases are the same as that of 2D cases. However, because the scale of the input 3D point cloud may vary, we first developed a method to obtain the scale $s$ to improve the adaptivity of the proposed methods, that is

\begin{equation}
s = \sqrt {\frac{1}{{2N}}\left( {\sum\limits_{i = 1}^N {{{({{\bf{x}}_i} - {\bf{\bar x}})}^2}}  + \sum\limits_{i = 1}^N {{{({{\bf{y}}_i} - {\bf{\bar y}})}^2}} } \right)},
\end{equation}

\noindent where $\bar {\bf{x}}$ and $\bar {\bf{y}}$ are the mean coordinates of ${\bf{x}}_i$ and ${\bf{y}}_i$, $i=1,2,...N$, respectively. Certain parameters are adjusted for the 3D case as follows: $H = 0.1 s$, $r = 0.3 s$ and $a = 20 / s$. Instead of normalizing the point cloud, we adjust the parameters according to scale $s$ to obtain the deformation field $f$ that can be directly used. In practice, we found that more neighboring matches are needed for 3D cases, hence we set $N_{\rm{neighbor}} = 50$.

\begin{figure*} [htp]
\vspace{0.0cm}
\centering
  \includegraphics[width=1.\textwidth]{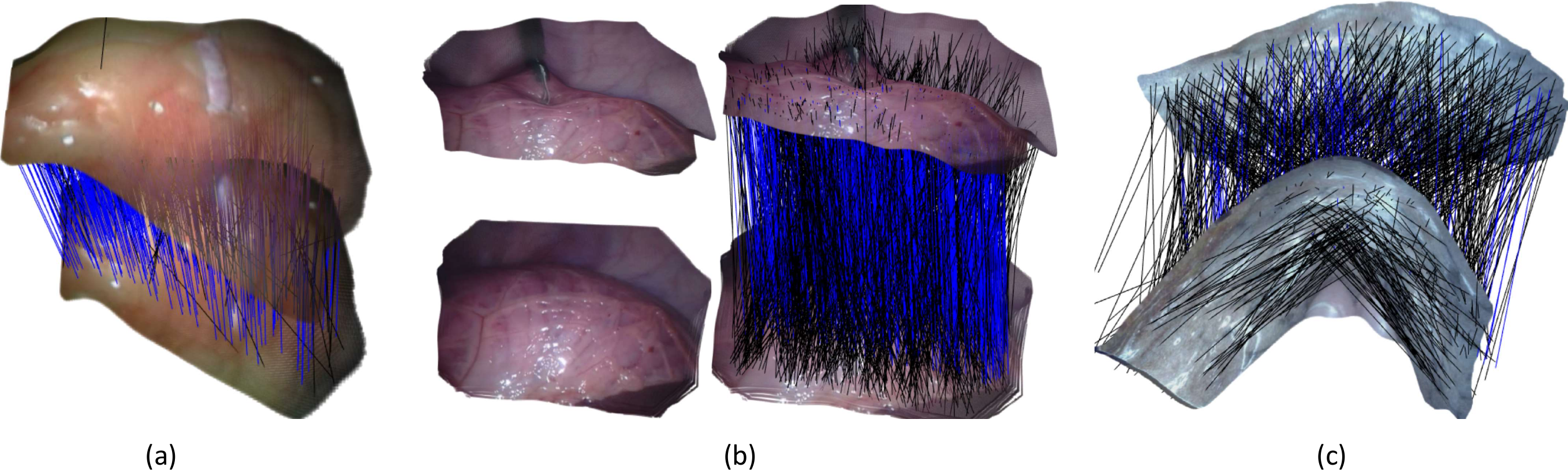}
\caption{3D experiments with soft tissue. The blue and black lines are identified as inliers and outliers by EMDQ respectively. (a) A silicon phantom that simulated heartbeat. (b) In vivo porcine abdomen, where the deformation was caused by instrument interaction. (c) A phantom with the texture of the lung surface, which has large deformation.}
\label{fig_3DMedicalImages}
\end{figure*}

We conducted the 3D experiments on medical data. As shown in Fig. \ref{fig_3DMedicalImages}, the 3D models were generated by a stereo matching method \cite{zhou2019real} from stereo laparoscopy videos. We performed ORB feature matching \cite{rublee2011orb} on the related 2D images and re-projected the feature points to the 3D models. The accuracy comparison results are in Table. \ref{tab_1}, which demonstrate that (s)EMDQ has the best accuracy. Compared with the results in Fig. \ref{fig_3DMedicalImages}, the $F$-$score$ is significantly higher in these experiments when the inliers ratio is low. This is because in these experiments, the modified ORB feature detection method can detect greater number of feature points than SURF, hence it is easier to generate the deformation field. The total number of matches are 629, 1784 and 693 for the three cases respectively, and the runtime of EMDQ are 53, 157 and 93 ms respectively.

\begin{table*}[]
\centering
\begin{tabular}{l|lll|lll|lll}
\multirow{2}{*}{} & \multicolumn{3}{c|}{case (a) (inliers ratio = 76\%)} & \multicolumn{3}{c|}{case (b) (inliers ratio = 39\%)} & \multicolumn{3}{c}{case (c) (inliers ratio = 16\%)}  \\ \cline{2-10}
                  & $recall$  & $precision$ & $F$-$score$       & $recall$ & $precision$ & $F$-$score$       & $recall$ & $precision$ & $F$-$score$       \\ \hline
LPM               & 0.93    & 0.91    & 0.92          & 0.91   & 0.92    & 0.91          & 0.90   & 0.92    & 0.91          \\
LMR               & 0.93    & 0.97    & 0.95          & 0.90   & 0.95    & 0.92          & 0.82   & 0.92    & 0.87          \\
SIM               & 0.97    & 0.67    & 0.79          & 0.98   & 0.43    & 0.60          & 0.97   & 0.32    & 0.48          \\
VFC/sVFC          & 0.99    & 0.94    & 0.96          & 0.99   & 0.96    & 0.97          & 0.99   & 0.94    & 0.96          \\
R1P-RNSC           & 0.99    & 0.95    & 0.97          & 0.99   & 0.90    & 0.94          & 0.99   & 0.92    & 0.95          \\
EMDQ              & 0.96    & 0.98    & 0.97          & 1.00   & 0.96    & \textbf{0.98} & 0.97   & 0.99    & \textbf{0.98} \\
sEMDQ             & 0.97    & 0.98    & \textbf{0.98} & 0.99   & 0.96    & 0.97          & 0.97   & 0.98    & {0.98} \\
\end{tabular}
\caption{Accuracy results of the soft tissue experiments as shown Fig. \ref{fig_3DMedicalImages}.}
\label{tab_1}
\end{table*}

\section{Conclusion}
We proposed two novel algorithms for solving the matching outliers removal problem, which are called as R1P-RNSC and EMDQ. The two algorithms compensate for the drawbacks of each other. Specifically, R1P-RNSC is fast to extract hidden rigid transforms from matches with outliers, but it does not take into account smoothing information hence its accuracy is limited. EMDQ is based on dual quaternion-based interpolation to generate the smooth deformation field and have high accuracy, but it relies on the results of R1P-RNSC for initialization. The combination of R1P-RNSC and EMDQ achieves the best accuracy in terms of the total number of errors and $F$-$score$ compared with other state-of-the-art methods. The proposed method is among the fastest methods for 2D cases, and can achieve real-time performance for 3D cases.

The proposed algorithms can be considered as a method to generate dense deformation field from sparse matches with outliers, which suggests that the displacements of all pixels can be obtained from the sparse matches (if uncertainty is not considered). Hence, the algorithms have potential to be used in many applications, such as non-rigid image registration. However, because the deformation field is generated by interpolating among the sparse feature matches,  the uncertainty problem needs to be further studied.

\appendices

\ifCLASSOPTIONcompsoc
  \section*{Acknowledgments}
\else
  \section*{Acknowledgment}
\fi

This work was supported by the National Institute of Biomedical Imaging and Bioengineering of the National Institutes of Health through Grant Numbers K99EB027177, R01EB025964, R01DK119269, P41EB015898, and a Research Grant from Siemens-Healthineers USA. Unrelated to this paper, Jayender Jagadeesan owns equity in Navigation Sciences, Inc. He is a co-inventor of a navigation device to assist surgeons in tumor excision that is licensed to Navigation Sciences.  Dr. Jagadeesan interests were reviewed and are managed by BWH and Partners HealthCare in accordance with their conflict of interest policies.

\ifCLASSOPTIONcaptionsoff
  \newpage
\fi

\bibliographystyle{IEEEtran}

\bibliography{IEEEabrv,bare_jrnl_compsoc}

\end{document}